\documentclass[11pt,a4paper]{article}
\usepackage[hyperref]{rocling2026}
\usepackage{times}
\usepackage{latexsym}

\usepackage{amsmath}
\usepackage{amssymb}
\usepackage{graphicx}
\usepackage[table]{xcolor}
\usepackage{booktabs}
\usepackage{simpleicons}

\usepackage{microtype}
\usepackage{fontawesome6}

\roclingfinalcopy 

\title{\textit{Fact over Fiction:} Detection of Pathological Hallucinations in Sinhala-to-English Neural Machine Translation}

\author{Navam Obeysekara$^\spadesuit$ \and Nevidu Jayatilleke$^\clubsuit$ \\
  $^\spadesuit$School of Computing, 
  Informatics Institute of Technology, 
  Sri Lanka \\
  $^\clubsuit$Department of Computer Science \& Engineering, University of Moratuwa, Sri Lanka \\
  \texttt{navam.20211058@iit.ac.lk, nevidu.25@cse.mrt.ac.lk}
}

\date{}

\begin{document}
\maketitle

\begin{abstract}
Neural Machine Translation (NMT) models, while capable of producing highly fluent outputs, remain vulnerable to hallucinations, which are translations that are natural yet semantically unrelated to the source. This vulnerability is acute in low-resource settings like Sinhala-to-English, where weak cross-lingual alignment leads to hallucinations. This
paper introduces a framework for reference-free hallucination detection in this language pair. We present a 45,000-sample synthetic dataset generated through a probabilistic chain of five linguistically motivated corruption strategies, with a semantic rescue mechanism that uses character-level similarity to distinguish hallucinations from morphological variants. We fine-tune \texttt{mDeBERTa-v3} for token-level sequence labelling, reaching a token-level F1 of $0.841 \pm 0.001$ over three seeds on a source-disjoint test set, and study a three-signal ensemble integrating neural risk scores, sequence log-probabilities, and cross-lingual semantic embeddings (\texttt{LaBSE}). A source-ablation control shows that the detector relies on the Sinhala source rather than on surface artefacts of the corruption process: shuffling or removing the source reduces sentence-level AUROC from 0.970 to chance. We benchmark eight NMT systems spanning five model families and find that detector firings vary by an order of magnitude across architectures.
\end{abstract}

\begin{keywords}
Low-Resource NLP, Hallucinations Detection, Machine Translation
\end{keywords}

\section{Introduction}
\label{sec:intro}

While the Transformer architecture \citep{vaswani2017attention} has established Neural Machine Translation (NMT) as the gold standard for cross-lingual communication, models remain critically vulnerable to hallucinations. These are fluent falsehoods that deceive users by mimicking valid output while fabricating information entirely \citep{ji2023survey, lee2018hallucinations, guerreiro-etal-2023-looking}.

This risk is magnified in Low-Resource Languages (LRLs), where the scarcity of high-quality parallel corpora leads to model uncertainty and exposure bias \citep{wang-sennrich-2020-exposure}. The reverse translation direction presents unique challenges. When translating from a low-resource source (Sinhala) to a high-resource target (English), the model's strong English language model can generate plausible-sounding text bearing no relation to the original content. While existing literature has predominantly focused on English-to-Sinhala
translation \citep{de2019survey}, the reverse direction remains severely
under-explored.

A complicating factor is Sinhala's severe diglossia. Real-world inputs frequently contain colloquial syntax, including sections that routinely omit subjects or pronouns absent from formal training corpora. Encountering this input, models may resort to \textit{syntactic forcing} by inserting grammatically required English agents or auxiliary verbs that have no direct source counterpart. Currently, existing evaluation benchmarks such as HalOmi \citep{dale-etal-2023-halomi} exclude the Sinhala-English pair entirely, leaving no labelled training data, evaluation framework, or detection baseline.

To address this, this paper presents a linguistically grounded study on detecting hallucinations in Sinhala-to-English NMT, arguing that effective detection in LRLs requires moving beyond standard uncertainty metrics. We introduce a synthetic dataset of 45,000 sentence pairs generated through five linguistically motivated corruption strategies with a semantic rescue mechanism. Using this data, we train a reference-free token-level \texttt{mDeBERTa-v3} detector (F1\,=\,$0.841 \pm 0.001$ over three seeds, on a test set sharing no source sentence with training). Because every synthetic corruption is a monolingual English edit, we add a source-ablation control that tests whether the detector actually uses the Sinhala source. We further study a three-signal ensemble combining neural detection, sequence log-probability, and \texttt{LaBSE} cross-lingual semantic similarity, and establish a hallucination benchmark across eight Sinhala-to-English NMT systems spanning five model families.

\section{Related Work}
\label{sec:related}

In this section, we review synthetic hallucination data, critiques of its realism, and detection strategies based on confidence, quality estimation, and cross-lingual embeddings.

\subsection{Synthetic Data Approaches}
A significant challenge in hallucination detection is the scarcity of naturally occurring labelled errors. \citet{zhou-etal-2021-detecting} pioneered the use of synthetic hallucination datasets by framing detection as a sequence labelling task. Their approach involved corrupting high-quality source sentences and reconstructing them via a pre-trained \texttt{BART} denoising autoencoder to produce fluent but semantically incorrect outputs \citep{lewis-etal-2020-bart}. They bridged the bilingual gap with bilingual lexicon induction \citep{zhou-etal-2019-density} and alignment heuristics such as \texttt{SimAlign} \citep{jalili-sabet-etal-2020-simalign}, enabling supervised token-level detectors such as \texttt{XLM-R} \citep{conneau-etal-2020-unsupervised} where no natural labels existed.

\subsection{Natural Hallucinations and Critique}
\citet{guerreiro-etal-2023-looking} argued that synthetic errors fundamentally differ from actual NMT hallucinations, introducing a taxonomy of oscillatory, strongly detached, and fully detached types. Their findings revealed that simple model confidence, measured via sequence log-probability, often performs on par with complex reference-based metrics while being significantly easier to implement. Furthermore, they proposed the DEHALLUCINATOR, which attempts to rewrite flagged translations using Monte Carlo dropout \citep{gal2016dropout}. This aligns with the hypothesis suggested by \citet{raunak-etal-2021-curious}, which discusses that hallucinations often stem from model instability or corpus noise rather than fundamental architectural flaws.

\subsection{Extended Detection Strategies}
The field has explored diverse statistical and glass-box strategies beyond synthetic methods. \citet{martindale-etal-2019-identifying} combined fluency scores with Bag-of-Vectors Sentence Similarity to identify translations that are fluent but inadequate, proving semantic similarity more reliable than BLEU \citep{papineni-etal-2002-bleu}. \citet{fomicheva2020unsupervised} evaluate Monte Carlo dropout as an
unsupervised QE indicator across six language pairs, with Sinhala-English
placed in their lowest-resource tier; they report weaker correlation with
human judgments for this pair relative to their higher-resource pairs
\citep{gal2016dropout, kim-etal-2017-predictor}. Parallel to this, \texttt{COMET} established a new paradigm for QE by regressing directly on human quality scores \citep{rei-etal-2020-comet}, while \texttt{ALTI+} measures the impact on generated tokens, hypothesising that low source contribution is a primary signal for hallucination \citep{ferrando-etal-2022-towards, dale-etal-2023-detecting}.

For cross-lingual embeddings, \citet{artetxe-schwenk-2019-massively} introduced \texttt{LASER}, which \citet{heffernan-etal-2022-bitext} extended via \texttt{LASER3} to scale detection to hundreds of languages. We adopt \texttt{LaBSE} \citep{feng-etal-2022-language}, a dual-encoder architecture optimised with Additive Margin Softmax loss to produce discriminative cross-lingual sentence embeddings. Despite these advances, sequence log-probability is often driven by the target-language prior rather than faithfulness to the source. In low-resource Sinhala-English settings, where the encoder provides a weak cross-lingual signal, the decoder can generate a high-probability English sentence while effectively ignoring the Sinhala source. This failure mode, characterised by confident but semantically untethered output, is precisely what motivates our reference-free neural approach.

A related concern for supervised detectors trained on synthetic data is that they may exploit artefacts of the data-construction process rather than the intended signal, as shown for hypothesis-only baselines in natural language inference \citep{poliak-etal-2018-hypothesis, gururangan-etal-2018-annotation}. We adopt the corresponding control (\S\ref{sec:ablation}).

\section{Methodology}
\label{sec:methodology}

The proposed framework\footnote{\url{https://github.com/Imanjith/synthetic-sinhala-english-dataset}}
operates as a two-stage pipeline consisting of a \textit{Generation} stage producing a
hypothesis and model confidence score, and a \textit{Detection} stage where a
fine-tuned cross-lingual encoder identifies hallucinated tokens. Formally,
let $S = \{s_1,\ldots,s_m\}$ be a Sinhala source and
$H = \{h_1,\ldots,h_n\}$ the English hypothesis. Hallucination detection
learns $f(S,H) \rightarrow Y$, where $Y = \{y_1,\ldots,y_n\}$ is a binary
label sequence over hypothesis tokens.

\subsection{Hallucination Taxonomy}

Following \citet{guerreiro-etal-2023-looking} and \citet{raunak-etal-2021-curious}, we
partition hallucinations into two categories. \textit{Detached hallucinations} occur when the model generates content that diverges from the factual meaning of the source on a severity spectrum, ranging from strongly detached outputs to fully detached outputs. Fully detached outputs have no semantic link whatsoever, whereas strongly detached outputs preserve the surface topic but fabricate specific facts. \textit{Fluent hallucinations} remain locally coherent but
introduce subtle semantic errors such as logic inversions and antonym substitutions, whilst remaining grammatically
well-formed. The mapping from categories to corruption strategies is summarised in Table~\ref{tab:taxonomy}.

\begin{table}[t]
\centering
\resizebox{0.9\columnwidth}{!}{
\begin{tabular}{l|l|l}
\hline
\textbf{Category} & \textbf{Strategy} & \textbf{Challenge} \\ \hline
Fully Detached    & High-Temp Sampling & Low conf. signal \\ \hline
Strongly Det.     & Entity (NER) Swap  & Fluent surface \\ \hline
Strongly Det.     & Numeric Distortion & Subtle factual error \\ \hline
Fluent            & Semantic Drift     & Same syntax \\ \hline
Fluent            & Dependency Swap    & Wrong roles \\ \hline
\end{tabular}}
\caption{Hallucination categories, corruption strategies, and detection challenges.}
\label{tab:taxonomy}
\end{table}

\subsection{Dataset Construction}
\label{sec:data}

We construct a synthetic benchmark\footnote{\href{https://huggingface.co/datasets/imanjith/synthetic-sinhala-english-hallucination-dataset}{\raisebox{-0.35em}{\includegraphics[height=1.5em]{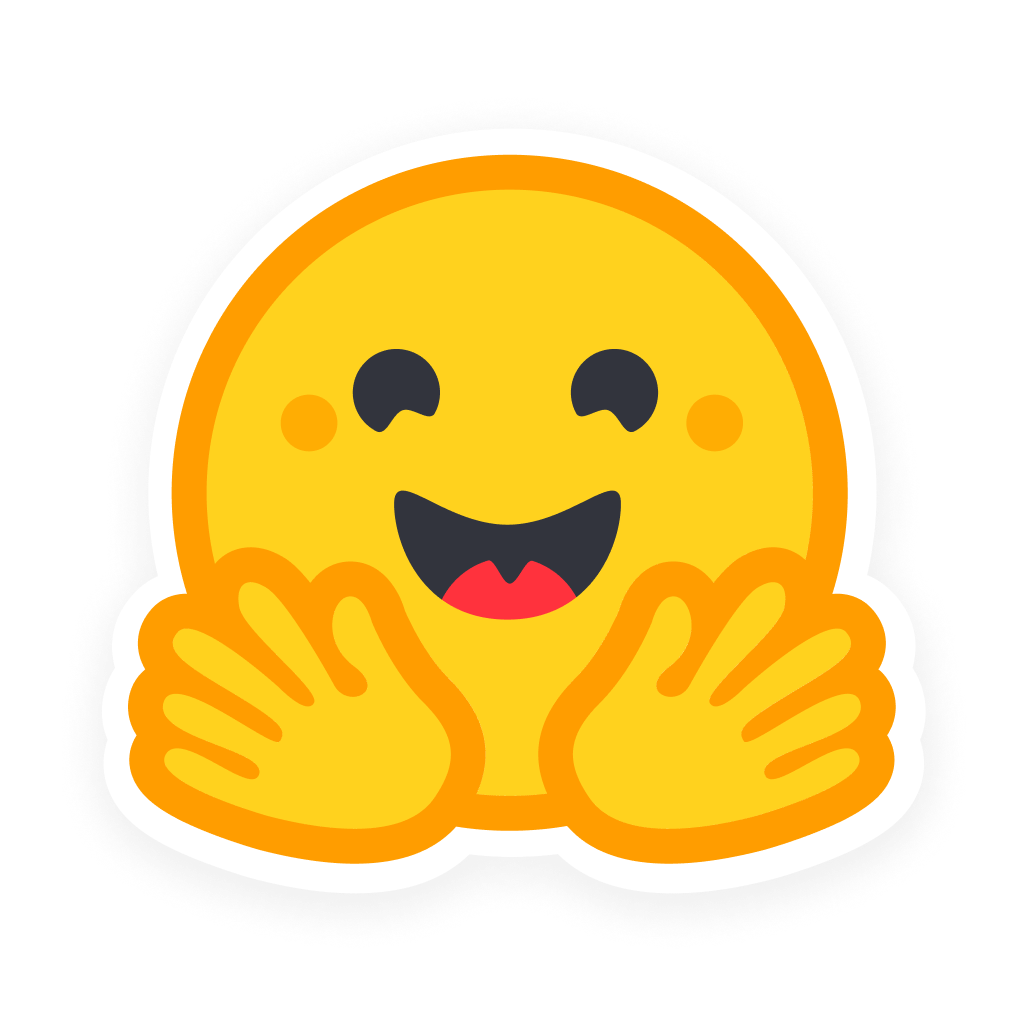}}~Synthetic-Sinhala-English-Hallucination-Dataset}} for hallucination detection in Sinhala-to-English translation. Starting from a clean parallel corpus, we corrupt English references with five strategies (alone or chained), filter the results for grammaticality, and release 45,000 labelled rows. A balanced 15,000-row subset, split by source sentence, is used for all detector experiments.

\subsubsection{Data Source and Preprocessing}

The dataset is built on the
\texttt{nllb-top25k-ensi-cleaned}\footnote{\small \url{https://huggingface.co/datasets/NLPC-UOM/nllb-top25k-ensi-cleaned}} corpus
\citep{ranathunga-etal-2024-quality}, a curated Sinhala-English parallel dataset of
approximately 25,000 sentence pairs. We removed 4 entries whose English text contained Sinhala characters or no Latin characters (0.02\%). From the remaining 24,996 pairs, 7,500 source sentences were sampled (seed 42), and five negative samples were planned per source, giving $7{,}500 \times (1+5) = 45{,}000$ rows. The pipeline is seeded and was run once; the released corpus is used in all experiments.

\subsubsection{Synthetic Hallucination Generation}

Negative samples were generated using a custom
\texttt{HallucinationGenerator} applying five corruption strategies. Deterministic strategies edit spaCy \citep{spacy} tokens by index, so that a replacement never hits a different occurrence or a substring of the target word. A shared entity pool (PERSON, GPE, ORG, DATE, etc.) was pre-built from 2,000 sampled reference sentences.

\paragraph{Entity Swapping (NER Swap).}
One named entity is replaced with a different entity of the same label
from the pre-built pool.

\paragraph{Semantic Drift.}
NLTK WordNet \citep{wordnet} replaces one non-entity content word with an antonym of the same part of speech. Candidates are adjectives and adverbs, or uninflected verbs and nouns (so that the antonym, a lemma, fits the context), excluding \textit{be}, \textit{have} and \textit{do}. Up to three candidates are tried; capitalisation of the original word is preserved.

\paragraph{Syntactic Confusion (Dependency Swap).}
spaCy's dependency parser identifies the syntactic subject (\texttt{nsubj})
and object (\texttt{dobj}/\texttt{obj}) of the same verb. When both are present and neither is a pronoun (pronoun swaps create trivially detectable case errors), their
positions are swapped to produce subject-object confusion (e.g., ``The manager
contacted the client'' $\rightarrow$ ``The client contacted the manager'').

\paragraph{Numeric Distortion.}
One whole number (thousands separators handled, decimals and alphanumeric tokens skipped) is incremented, decremented, or extended by appending a random digit. The increment/decrement magnitude is drawn from
$[1, \max(10, \lfloor 0.2v \rfloor)]$, where $v$ is the original value, ensuring distortion proportional to the
original magnitude.

\paragraph{High-Temperature Sampling.}
English ground-truth sentences are re-encoded through
\texttt{NLLB-200-distilled-1.3B} \citep{nllb2022} at $T\!=\!1.5$ with
nucleus sampling (top-p\,=\,0.95, top-k\,=\,50). A post-hoc filter retains
only outputs with BERTScore F1\,$<$\,0.92 \citep{zhang2019bertscore} (\texttt{roberta-large}, no baseline rescaling). The threshold was set by manual inspection of samples in the 0.88--0.94 range: below 0.92 outputs predominantly altered meaning, while 0.92--0.94 mixed genuine changes with close paraphrases. Of 31,768 sampled outputs, 11,388 (35.8\%) were retained.

\subsubsection{Probabilistic Chain Generation}

Rather than applying strategies independently, a probabilistic chain produces
compound hallucinations in three phases. \textit{Phase 1 (Planning):} each negative sample begins as High-Temperature Sampling
(p\,=\,0.8) or a deterministic strategy (p\,=\,0.2). \textit{Phase 2
(Batched GPU Generation):} high-temperature samples are generated in batches
of 32 and filtered by BERTScore; rejected samples fall back to a deterministic strategy. \textit{Phase 3 (Secondary
Mutation):} a second, distinct strategy is applied with p\,=\,0.8 (e.g.,
\texttt{temp+ner}). Whenever a deterministic strategy is needed, all applicable strategies are tried in random order until one changes the text; a row left identical to its reference is relabelled 0.

\begin{figure*}[t]
\centering
\includegraphics[width=0.95\textwidth]{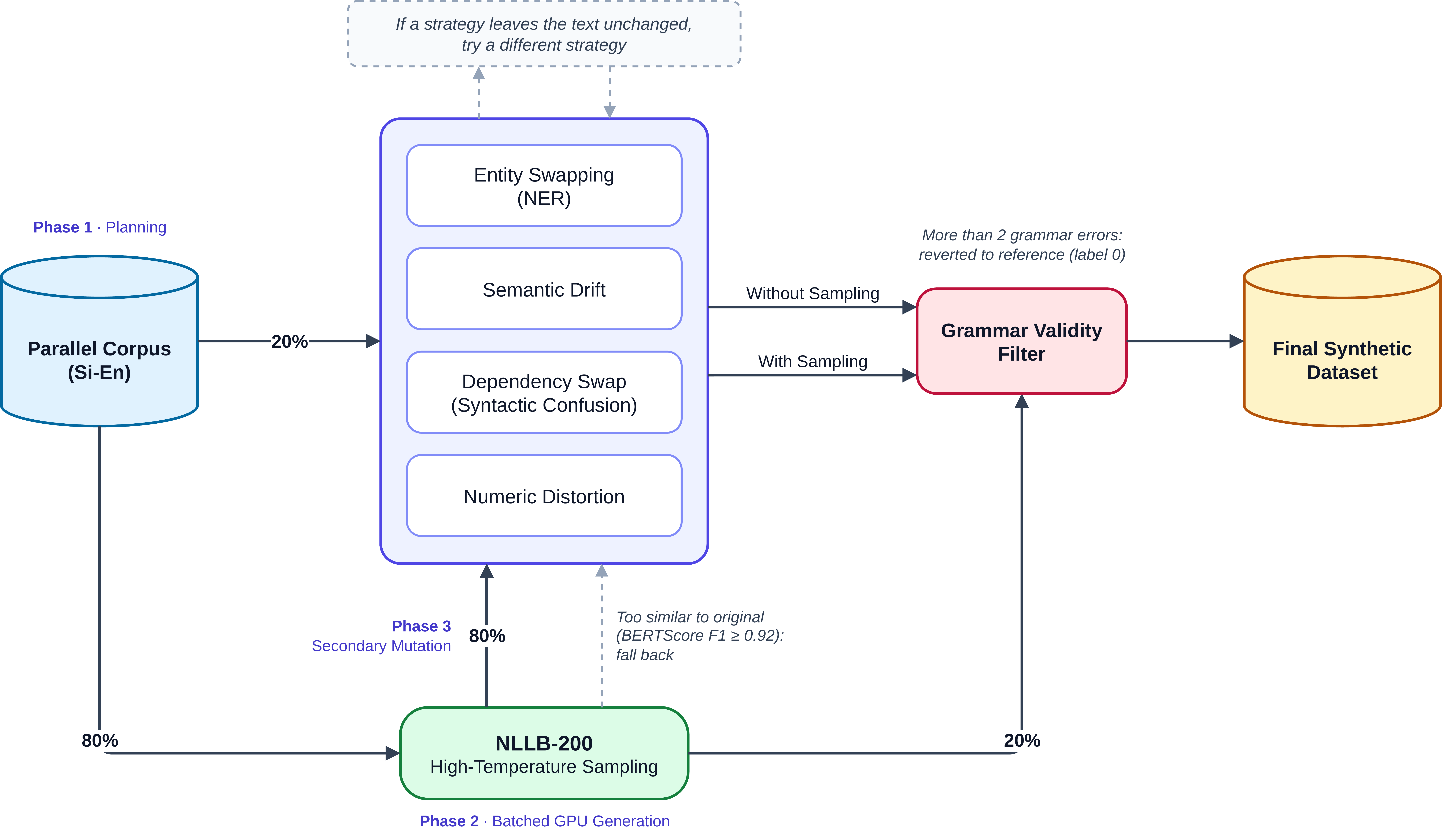}
\caption{Dataset generation pipeline showing the three-phase probabilistic chain.}
\label{fig:pipeline}
\end{figure*}

\subsubsection{Grammar Validity Filter}

Before balancing, all hallucinated rows are filtered using \texttt{language\_tool\_python}, ignoring
PUNCTUATION, TYPOGRAPHY, WHITESPACE, and CASING categories. Rows with more
than two remaining grammar errors are reverted to the reference and relabelled 0.
The filter rejected 109 of 32,686 hallucinated rows (0.33\%). It operates purely on grammaticality, so meaning-altering but syntactically disrupted samples are reverted too. The effect is small but not uniform across strategies: 78 of the 109 rejections involved High-Temperature Sampling, which is correspondingly under-represented after filtering. We did not manually verify the reverted samples.

\subsubsection{Corpus Composition and Splits}

The released corpus contains 45,000 rows: 32,577 hallucinated (label~1) and 12,423 faithful (label~0). The faithful rows comprise the 7,500 references and 4,923 reverted rows (4,811 rejected high-temperature samples with no applicable fallback, 109 grammar rejections, 3 failed fallbacks), which duplicate their references and are retained, marked by their \texttt{method} field, for transparency. The distribution by base (first-applied) strategy is given in Table~\ref{tab:dataset_stats} where 13,987 hallucinated rows (42.9\%) are compound.

For detector experiments we draw a balanced subset of 15,000 rows: the 7,500 references as label~0 and 7,500 randomly sampled hallucinated rows as label~1 (seed 42). This subset is split 70/10/20 by \emph{source sentence}, so that no Sinhala source appears in more than one partition, giving 10,486 training, 1,497 validation and 3,017 test rows (5,197 / 743 / 1,485 unique sources). Validation is used for checkpoint selection and threshold calibration; the test set is used only for final reporting.

\begin{table}[t]
  \centering
  \resizebox{0.9\columnwidth}{!}{
  \begin{tabular}{l|r|r}
    \toprule
    \textbf{Base strategy} & \textbf{Full corpus} & \textbf{Balanced} \\
    \midrule
    Ground truth (references) &  7,500 & 7,500 \\
    Reverted (faithful)       &  4,923 &     -- \\
    Semantic Drift            & 12,718 & 2,876 \\
    High-Temp Sampling        & 11,310 & 2,583 \\
    Entity (NER) Swap         &  6,717 & 1,596 \\
    Numeric Distortion        &    937 &   239 \\
    Dependency Swap           &    895 &   206 \\
    \midrule
    \textit{Total}            & \textit{45,000} & \textit{15,000} \\
    \bottomrule
  \end{tabular}}
  \caption{Corpus composition after grammar filtering, by base (first-applied) strategy; compound rows are counted under their first strategy.}
  \label{tab:dataset_stats}
\end{table}

\subsection{Token-Level Label Construction}
\label{sec:label_construction}

Token-level binary labels are derived from
\texttt{difflib.SequenceMatcher} alignment between corrupted hypothesis and
ground-truth reference. After light normalisation, the aligner assigns label~0
to \texttt{equal} and \texttt{delete} operations, and candidate label~1 to
\texttt{replace} and \texttt{insert} operations. Omissions therefore carry no label.

\paragraph{Semantic Rescue for Morphological Variants.}
High-temperature outputs frequently contain surface variants of reference words (e.g., ``colour'' vs.\ ``color'', ``believes'' vs.\ ``believed'') that should not count as hallucinations. For rows involving High-Temperature Sampling, each candidate label-1 word $w_h$ is compared against the words $A$ of the aligned reference span (extended by three words on either side) using the Ratcliff/Obershelp similarity implemented by \texttt{difflib}:
\begin{equation}
  \text{sim}(w_h, w_r) = \frac{2M}{|w_h| + |w_r|},
\end{equation}
where $M$ is the number of matching characters found by recursive longest-common-substring matching on lower-cased words. If $\max_{w_r \in A} \text{sim}(w_h, w_r) \geq \theta_r = 0.80$, the label is
downgraded to~0 (e.g., ``colour''/``color'': 0.91). Rows produced only by deterministic strategies are hallucinations by construction and receive no rescue. Restricting comparison to the aligned span prevents words that merely occur elsewhere in the reference, such as the two arguments of a dependency swap, from being rescued. Word-level labels are propagated to all sub-word tokens of the word, while source and special tokens receive label~$-100$.

Rescue does not detect lexical paraphrases between surface-dissimilar words (e.g, ``stated'' vs.\ ``said'', similarity 0.60; ``perception'' vs.\ ``understanding''). Such words receive label~1 in high-temperature rows even when meaning is preserved, which is a source of label noise and of false positives at deployment.

\textbf{Sensitivity to $\theta_r$.} In the 2,583 high-temperature rows of the balanced subset, alignment yields 32,889 candidate label-1 words. The share rescued is 7.7\% at $\theta_r = 1.0$, 8.7\% at 0.90, 13.3\% at 0.80 and 15.6\% at 0.70. Of 20 random rescued words per band, all 20 in [0.90, 1.0) were surface variants, 11 in [0.80, 0.90), and only 3 in [0.70, 0.80), where the rest were different words (e.g., ``their''/``the''). We keep $\theta_r = 0.80$ as the point below which rescues are mostly errors, with short words the main source of false rescues. This analysis concerns the labels only; we did not retrain the detector at other thresholds.

\subsection{Translation and Confidence Scoring}

The sequence-level confidence of a generated translation is its log-probability under the generating model,
\begin{equation}
    \text{LogProb}(H|S) = \sum_{t=1}^{|H|} \log p(h_t \mid h_{<t}, S),
\end{equation}
summed over the transition scores of the beam-selected hypothesis as returned by beam search, so that the score is exactly the one the decoder used to select $H$ (we verified that per-token averages match the reported beam scores for all eight systems). For the synthetic evaluation in \S\ref{sec:ensemble_eval}, where the hypothesis is given rather than generated, the same quantity is computed by teacher forcing under \texttt{NLLB-200-distilled-1.3B}. Scores are min-max normalised per model (\S\ref{sec:ensemble}).

\subsection{Hallucination Detection Architecture}
\label{sec:detection}

\texttt{mDeBERTa-v3} (\texttt{mdeberta-v3-base})
\citep{he2021debertav3} is selected as the primary architecture for two
principled reasons. First, its Replaced Token Detection (RTD) pre-training is
structurally identical to our task. The model learns to classify each token as
original or replaced, which is a discriminative objective that makes it
${\approx}$6.7$\times$ more sample-efficient than masked language modelling
and whose training signal directly mirrors the hallucination detection setting.
Second, its disentangled attention maintains separate matrices for content and
relative position, which is advantageous for the cross-encoder input format
$X = \texttt{[CLS]} \oplus S \oplus \texttt{[SEP]} \oplus H \oplus
\texttt{[SEP]}$. In the concatenated source-hypothesis sequence, positional
proximity between a Sinhala source token and an English hypothesis token
carries no semantic information, as the two segments are separated by a
boundary and have different word-order typologies. Disentangled attention
allows content-to-content interactions to be weighted independently of this
positional distance. On XNLI zero-shot evaluation, \texttt{mDeBERTa-v3} outperforms
\texttt{XLM-RoBERTa} by 3.6\% \citep{he2021debertav3}, which serves as our baseline.

\subsubsection{Input Representation and Training}

The hypothesis is tokenised word by word so that labels align with words; the source is truncated first if the pair exceeds 512 tokens. Each hypothesis token hidden state $h_i$ ($d=768$) is classified by a two-class softmax head,
\begin{equation}
    P(y_i \mid X) = \text{softmax}(W h_i + b),
\end{equation}
optimised with cross-entropy. Configuration: $lr=2\times10^{-5}$,
6~epochs, batch~size~8, weight decay~0.01, fp16, single NVIDIA T4. The checkpoint with the highest validation token F1 is kept. Each detector is trained with three seeds (42, 43, 44); the same code, data and hyperparameters are used for the \texttt{XLM-RoBERTa} baseline.

\subsubsection{Inference and Risk Scoring}

A token is flagged if $P(y_i=1 \mid X) > 0.5$. The \textit{Hallucination Ratio} aggregates these into a sentence-level score:
\begin{equation}
    R_H = \frac{\sum_{i=1}^{n} \mathbb{I}(\hat{y}_i = 1)}{n}
\end{equation}
and a sentence is flagged if $R_H > \tau$. We calibrate $\tau$ on the validation set by sentence-level F1 over $\tau \in \{0, 0.01, \ldots, 0.04, 0.05, 0.10, \ldots, 0.50\}$. All three seeds select $\tau=0$ (validation F1 0.950--0.961), so the calibrated rule flags a sentence whenever \emph{any} hypothesis token is flagged. This removes the insensitivity of ratio thresholds to single-token hallucinations in long sentences, which matters for domains such as legal or medical translation where a single wrong number or entity is consequential. We also report the maximum token probability as an alternative, threshold-free sentence score.

\subsection{Three-Signal Ensemble Framework}
\label{sec:ensemble}

The neural detector $R_H$, inverted log-probability $\hat{C}$, and \texttt{LaBSE}
similarity $\hat{\rho}$ capture different aspects of translation
quality. Spearman correlations between log-probability and $R_H$ range from
$\rho=-0.103$ to $\rho=-0.437$ across the eight benchmarked systems (Table~\ref{tab:spearman}), confirming the signals are related but
non-redundant. Their ensemble score is:
\begin{equation}
    E = w_{\text{det}} \cdot R_H + w_{\text{lp}} \cdot \hat{C} +
        w_{\text{sem}} \cdot (1 - \hat{\rho})
    \label{eq:ensemble}
\end{equation}
and a sentence is flagged if $E$ exceeds a threshold. We evaluate the originally proposed weighting ($w_{\text{det}}=0.5$, $w_{\text{lp}}=0.3$, $w_{\text{sem}}=0.2$) alongside four alternative configurations and a weighting tuned on validation; each configuration's threshold is tuned on validation (\S\ref{sec:ensemble_eval}).

\paragraph{\texttt{LaBSE} Semantic Similarity.}
\texttt{LaBSE} \citep{feng-etal-2022-language} encodes source and hypothesis via
mean-pooled, L2-normalised representations in a shared multilingual space.
Cosine similarity below 0.80 is treated as semantic divergence, independent of
reference translations.

\paragraph{Log-Probability Normalisation.}
\begin{equation}
    \hat{C} = \frac{\min(C_{\text{raw}}, 0) - C_{\text{max}}}{C_{\text{min}} - C_{\text{max}}}
\end{equation}
where higher $\hat{C}$ indicates lower confidence and $C_{\text{min}}, C_{\text{max}}$ are taken per model (on the benchmark) or from the validation set (on the synthetic evaluation).

\subsection{Evaluation Protocol}
\label{sec:benchmarking_method}

\paragraph{Detector.} Token-level precision, recall and F1 over hypothesis tokens of the source-disjoint test split, reported as mean $\pm$ standard deviation over three seeds, overall and by base strategy.

\paragraph{Source-ablation control.} Because every corruption is a monolingual English edit of the reference, a detector could learn to recognise perturbed English without consulting the source. We therefore evaluate the trained detectors with (i) the original source, (ii) the source replaced by a different, length-matched test source, and (iii) the source removed, keeping the gold labels fixed; (iv) retrain the detector with the source removed from training, validation and test (a hypothesis-only baseline); and (v) pair \emph{faithful} test hypotheses with a mismatched source. Case (v) creates fully detached pairs whose English is untouched: a source-aware detector must flag them, whereas a detector relying on perturbation artefacts will not.

\paragraph{Ensemble.} On the synthetic validation and test splits, where the gold label refers to the scored hypothesis, we compute $R_H$ (seed-42 detector), teacher-forced log-probability and \texttt{LaBSE} similarity. Weights (simplex grid, step 0.1) and thresholds are tuned on validation and reported on test, together with per-signal AUROC.

\paragraph{NMT benchmark.} Eight Sinhala-to-English systems translate the 1,485 unique test sources: \texttt{NLLB-200} (distilled 600M and 1.3B) \citep{nllb2022}, \texttt{M2M-100}
(418M and 1.2B) \citep{fan2021beyond}, \texttt{mBART-50} (many-to-many and
many-to-one) \citep{liu-etal-2020-multilingual}, a community fine-tune of \texttt{mT5} for Sinhala-English (\texttt{mt5-sinhalese-english}\footnote{\url{https://huggingface.co/thilina/mt5-sinhalese-english}}) \citep{xue-etal-2021-mt5}, and
\texttt{Helsinki-NLP OPUS-MT} (mul-en). All use beam search with
\texttt{num\_beams}\,=\,4 and \texttt{max\_new\_tokens}\,=\,256; all translations completed successfully. We record which signals fire per sentence and the Spearman correlation between log-probability and $R_H$.

\section{Evaluation}
\label{sec:evaluation}

We first evaluate the detector on the synthetic test split, then verify with a source-ablation control that it relies on the source sentence. Next we assess the ensemble and apply it to benchmark eight NMT systems on 1,485 held-out test sources, and finally examine sources of false positives.

\subsection{Detector Performance}

\texttt{mDeBERTa-v3} achieves F1 of $0.841 \pm 0.001$,
outperforming \texttt{XLM-RoBERTa} by 5.2 points. The gap is largest in recall (+7.5 points), consistent with RTD pre-training, which trains the model to identify substituted tokens \citep{he2021debertav3}. Token-level test performance is presented in Table~\ref{tab:detector_metrics}.

\begin{table}[t]
  \centering
  \resizebox{\columnwidth}{!}{%
  \begin{tabular}{l|c|c|c|c}
    \toprule
    \textbf{Model} & \textbf{Acc.} & \textbf{Prec.} & \textbf{Rec.} & \textbf{F1} \\
    \midrule
    \texttt{mDeBERTa-v3} & 0.946$\pm$0.001 & 0.845$\pm$0.014 & 0.837$\pm$0.011 & 0.841$\pm$0.001 \\
    \texttt{XLM-R}       & 0.930$\pm$0.001 & 0.819$\pm$0.023 & 0.762$\pm$0.028 & 0.789$\pm$0.005 \\
    \bottomrule
  \end{tabular}%
  }
  \caption{Token-level detector performance on the source-disjoint test split (3,017 sentences, hypothesis tokens only), mean $\pm$ std over three seeds.}
  \label{tab:detector_metrics}
\end{table}

Per-strategy F1 shows \texttt{mDeBERTa}'s advantage is consistent across all
strategies. Dependency Swap is the hardest type (\texttt{mDeBERTa}: 0.804 vs.\ \texttt{XLM-R}: 0.752), as the swapped words are all present in the source and only their roles change; High-Temperature Sampling follows (0.828 vs.\ 0.784), reflecting its paraphrase-related label noise. Semantic Drift (0.934 vs.\ 0.857), Numeric Distortion (0.933 vs.\ 0.855) and Entity Swap (0.916 vs.\ 0.860) are easiest, as single-word substitutions contradict an identifiable source counterpart.

\subsection{Source-Ablation Control}
\label{sec:ablation}

Replacing or removing the source at test time reduces token F1 from 0.841 to about 0.30 and sentence AUROC from 0.970 to chance (0.54 and 0.51), and the detector then flags every test sentence. Faithful hypotheses are flagged 4.5\% of the time with their own source but 100\% with a mismatched one, even though their English is unchanged. The detector therefore depends on the source-hypothesis relationship, not the hypothesis alone. The control results are reported in Table~\ref{tab:source_ablation}.

The hypothesis-only baseline reaches token F1 0.622 and AUROC 0.850, showing that the synthetic corruptions do leave monolingual cues (e.g., antonym substitutions and disfluent high-temperature outputs). The full model exceeds it by 0.22 F1 and 0.12 AUROC, and can also catch fluent but fully detached translations.

\begin{table}[t]
  \centering
  \resizebox{0.95\columnwidth}{!}{
  \begin{tabular}{l|c|c}
    \toprule
    \textbf{Condition} & \textbf{Token F1} & \textbf{AUROC} \\
    \midrule
    Original source        & 0.841$\pm$0.002 & 0.970$\pm$0.002 \\
    Shuffled source        & 0.314$\pm$0.003 & 0.540$\pm$0.007 \\
    Source removed         & 0.295$\pm$0.004 & 0.507$\pm$0.028 \\
    Hypothesis-only model  & 0.622$\pm$0.008 & 0.850$\pm$0.010 \\
    \bottomrule
  \end{tabular}}
  \caption{Source-ablation control on the test split (mean $\pm$ std over three seeds; AUROC of $R_H$). Gold labels are unchanged across conditions. Faithful hypotheses are flagged 4.5\%\,$\pm$\,0.9 of the time with their own source and 100\% with a mismatched source.}
  \label{tab:source_ablation}
\end{table}

\subsection{Ensemble Evaluation}
\label{sec:ensemble_eval}

 As single signals, the detector ($R_H$) reaches AUROC 0.972, well above inverted log-probability (0.880) and \texttt{LaBSE} (0.882). The detector alone attains F1 0.963 at its validation-tuned threshold. The originally proposed weighting (Det-Mild) is lower (0.927), and configurations emphasising log-probability or \texttt{LaBSE} are lower still (0.871--0.907). The validation-tuned weighting assigns zero weight to log-probability and 0.1 to \texttt{LaBSE}; it matches the detector's F1 and slightly improves ranking (AUROC 0.979). On synthetic data, therefore, the detector carries nearly all of the signal, \texttt{LaBSE} adds a small complementary gain, and log-probability adds none. Ensemble results on the synthetic test split are reported in Table~\ref{tab:weight_ablation}.

\begin{table}[t]
  \centering
  \small
  \setlength{\tabcolsep}{3pt}
  \resizebox{\columnwidth}{!}{%
  \begin{tabular}{l|c|c|c|c|c|c}
    \toprule
    \textbf{Config} & \textbf{Weights} & \textbf{Prec.} & \textbf{Rec.} & \textbf{F1} & \textbf{Flag\%} & \textbf{AUROC} \\
    \midrule
    Detector only & 1/0/0       & 0.949 & 0.977 & 0.963 & 51.7 & 0.972 \\
    Det-Mild      & 0.5/0.3/0.2    & 0.916 & 0.938 & 0.927 & 51.5 & 0.971 \\
    Det-Heavy     & 0.7/0.2/0.1    & 0.941 & 0.962 & 0.951 & 51.4 & 0.977 \\
    LP-Heavy      & 0.2/0.6/0.2    & 0.812 & 0.940 & 0.872 & 58.1 & 0.938 \\
    Sem-Heavy     & 0.2/0.2/0.6    & 0.838 & 0.907 & 0.871 & 54.4 & 0.944 \\
    Uniform       & 0.33/0.33/0.33 & 0.879 & 0.938 & 0.907 & 53.6 & 0.961 \\
    Val-tuned     & 0.9/0/0.1 & 0.949 & 0.977 & 0.963 & 51.7 & 0.979 \\
    \bottomrule
  \end{tabular}%
  }
  \caption{Ensemble configurations on the synthetic test split (seed-42 detector; 50.2\% hallucinated). Weights are $w_{det}/w_{lp}/w_{sem}$; weights and thresholds tuned on validation.}
  \label{tab:weight_ablation}
\end{table}

\subsection{NMT Benchmarking Results}
\label{sec:benchmarking}

Each system's 1,485 translations are partitioned by which of the three signals \textit{fire} (Table~\ref{tab:model_benchmarking}): the detector fires when any hypothesis token is flagged ($\tau=0$), log-probability fires when a sentence falls below the model's own 30th percentile, and \texttt{LaBSE} fires when cosine similarity falls below 0.80.

\begin{table*}[t]
  \centering
  \resizebox{\textwidth}{!}{%
  \begin{tabular}{l|l|r|r|r|r|r|r|r|r|r|r}
    \toprule
    \textbf{Model} & \textbf{Family} & \textbf{LP/word} & \textbf{Det fires} &
    \textbf{All 3} & \textbf{Det+LP} & \textbf{Det+\texttt{LaBSE}} & \textbf{LP+\texttt{LaBSE}} &
    \textbf{Det Only} & \textbf{LP Only} & \textbf{\texttt{LaBSE} Only} & \textbf{None} \\
    \midrule
    \texttt{NLLB 600M}    & NLLB   & $-0.671$ &  51 &   6 &  22 &  1 &  6 & 22 & 412 & 12 & 1004 \\
    \texttt{NLLB 1.3B}    & NLLB   & $-0.617$ &  70 &   7 &  34 &  6 & 11 & 23 & 394 & 23 &  987 \\
    \texttt{M2M 418M}     & M2M    & $-0.919$ & 253 &  38 & 114 & 25 & 29 & 76 & 265 & 56 &  882 \\
    \texttt{M2M 1.2B}     & M2M    & $-0.815$ & 129 &  23 &  53 & 13 & 32 & 40 & 338 & 49 &  937 \\
    \texttt{mBART50 m2m}  & mBART  & $-0.734$ & 140 &  14 &  73 & 13 &  8 & 40 & 351 & 31 &  955 \\
    \texttt{mBART50 m2o}  & mBART  & $-0.707$ &  93 &   5 &  48 & 10 &  9 & 30 & 384 & 26 &  973 \\
    \texttt{mT5 Sin}      & mT5    & $-0.971$ & 329 & 109 & 113 & 49 & 29 & 58 & 195 & 61 &  871 \\
    \texttt{OPUS-mul-en}  & Marian & $-0.865$ & 433 & 134 & 103 & 99 & 26 & 97 & 183 & 67 &  776 \\
    \bottomrule
  \end{tabular}%
  }
  \caption{NMT benchmarking on 1,485 unique test sources per system. Partition columns (All 3 to None) sum to 1,485; \textbf{Det fires} is the total number of detector-flagged sentences; \textbf{LP/word} is mean sequence log-probability per output word.}
  \label{tab:model_benchmarking}
\end{table*}

\paragraph{Detection rates vary by architecture.}
These counts are detector firings rather than verified hallucinations (\S\ref{sec:fp}), so they indicate relative risk rather than error rates. Detector firings vary by nearly an order of magnitude, from 51 (\texttt{NLLB 600M}) to 433 (\texttt{OPUS-MT}). \texttt{OPUS-MT} and the \texttt{mT5} fine-tune produce the most detections and the most all-three joint detections (134 and 109), and they are also among the least confident systems per word. The \texttt{NLLB} models produce the fewest detections and the highest per-word confidence, consistent with their strong reported FLORES-200 performance on Sinhala \citep{nllb2022}. Scaling \texttt{M2M-100} from 418M to 1.2B roughly halves detections (253 to 129), whereas the two \texttt{NLLB} capacities perform similarly (51 and 70), and the many-to-one \texttt{mBART-50} variant produces fewer detections than the many-to-many variant (93 vs.\ 140). These patterns suggest family and training data matter more than parameter count here, though confirming this requires controlled experiments.

\paragraph{No single signal is sufficient.}
Det Only sentences (22--97 per system) are detector flags with normal confidence and high \texttt{LaBSE} similarity, which a confidence-only system would miss. Log-probability alone flags about 30\% of outputs by construction, producing 183--412 LP-Only sentences, most of which are uncertain but not flagged by the other signals. \texttt{LaBSE}-only detections are 12--67 per system, consistent with a confirmatory role.

\paragraph{Signal independence.}
All are negative and significant ($p<0.001$). Agreement is strongest for the weakest systems (\texttt{mT5} $-0.437$, \texttt{OPUS-MT} $-0.369$) and weakest for \texttt{NLLB} ($-0.103$, $-0.132$): for strong systems, the detections that remain are largely ones the model itself is confident about. Spearman correlations between log-probability and $R_H$ are reported in Table~\ref{tab:spearman}. Even the strongest correlation accounts for only ${\approx}$19\% of the variance in rank, confirming non-redundancy. 

\begin{table}[t]
\centering
\resizebox{0.6\columnwidth}{!}{%
\begin{tabular}{l|c}
\toprule
\textbf{Model} & \textbf{Spearman $\rho$} \\
\midrule
\texttt{NLLB 600M}    & $-0.103$  \\
\texttt{NLLB 1.3B}    & $-0.132$  \\
\texttt{M2M 418M}     & $-0.303$  \\
\texttt{M2M 1.2B}     & $-0.215$  \\
\texttt{mBART50 m2m}  & $-0.248$  \\
\texttt{mBART50 m2o}  & $-0.176$  \\
\texttt{mT5 Sin}      & $-0.437$  \\
\texttt{OPUS-mul-en}  & $-0.369$  \\
\bottomrule
\end{tabular}}
\caption{Spearman $\rho$ between log-probability and $R_H$ on the benchmark (all $p<0.001$).}
\label{tab:spearman}
\end{table}

\paragraph{Threshold sensitivity.}
Validation F1 is stable for $\tau \le 0.03$ (0.960--0.961) and declines monotonically thereafter (0.931 at 0.05, 0.571 at 0.20, 0.377 at 0.50), so ratio thresholds such as the commonly used 0.20 substantially under-flag. On the benchmark, detector firings decrease monotonically with $\tau$ for every system, and \texttt{OPUS-MT} and \texttt{mT5} remain the two highest-flagging systems at every threshold tested; ordering among the low-count systems varies.

\subsection{Sources of False Positives}
\label{sec:fp}

Without human annotation of the benchmark outputs we cannot measure precision on natural translations. As an indication, of the 129 \texttt{M2M 1.2B} sentences flagged by the detector, 36 (27.9\%) also fall below the \texttt{LaBSE} threshold; the remainder are either fluent hallucinations that preserve overall sentence meaning or false positives. Inspection suggests two recurring sources of false positives.

\paragraph{Lexical paraphrasing.}
The detector does not reliably accept synonymous substitutions. This follows from the label construction (\S\ref{sec:label_construction}): surface-dissimilar synonyms such as ``stated''/``said'' are labelled as hallucinations in high-temperature training rows, and semantic rescue cannot recover them.

\paragraph{Structural variations.}
When colloquial Sinhala input is encountered, NMT systems insert grammatically required English agents or auxiliaries with no direct source counterpart (\textit{syntactic forcing}). Word-level alignment against the reference penalises such insertions during training, so the detector can flag valid insertions as target-side hallucinations.

\section{Conclusion}
\label{sec:conclusion}

This work addresses hallucination detection in Sinhala-to-English NMT, a task with no prior labelled data, benchmarks, or baselines. We presented a 45,000-sample synthetic corpus, a reference-free token-level detector (\texttt{mDeBERTa-v3}, F1\,=\,$0.841 \pm 0.001$ on a source-disjoint test set), a source-ablation control showing that the detector depends on the Sinhala source, a three-signal ensemble evaluation, and a hallucination benchmark across eight NMT systems. Detector firings vary by an order of magnitude across architectures, and Det Only detections show that fluent, confident hallucinations exist that log-probability alone misses. On synthetic data the detector carries nearly all of the ensemble's signal, with \texttt{LaBSE} adding a small ranking gain.

Future work should prioritise human annotation of natural translation errors to measure real precision, and replace \texttt{difflib} supervision with cross-lingual word alignment to reduce paraphrase-induced label noise. Ensemble weights and thresholds should also be recalibrated on annotated natural translations, since values tuned on balanced synthetic data do not transfer to the much lower hallucination rates of real output.

\section*{Limitations}

The detector is trained and evaluated on synthetically corrupted data; generalisation to naturally occurring translation errors remains unverified. All corruptions are monolingual English edits of the reference, and High-Temperature Sampling re-encodes English rather than translating from Sinhala, so authentic Sinhala-origin failure modes may be under-represented. The source-ablation control (\S\ref{sec:ablation}) shows that the detector relies on the source, but the hypothesis-only baseline (F1 0.622) shows that the data also contain exploitable monolingual cues.

Token labels come from \texttt{difflib} alignment, which cannot recognise lexical paraphrases and ignores omissions. The rescue threshold $\theta_r$ was assessed through its effect on labels (\S\ref{sec:label_construction}) rather than by retraining at alternative thresholds, and at $\theta_r=0.80$ some short, different words are wrongly rescued. The high-temperature BERTScore threshold was chosen by qualitative inspection, so paraphrases below it are labelled as hallucinations. The grammar filter reverts samples on grammaticality alone and disproportionately affects high-temperature samples (78 of 109 rejections).

Detector results are averaged over three seeds, but the ensemble evaluation and NMT benchmark use a single detector (seed 42). The benchmark reports signal-firing counts without human ground truth, and the \texttt{LaBSE} agreement in \S\ref{sec:fp} is not a precision estimate. Because thresholds were calibrated on a balanced synthetic split, they over-flag real outputs, which is why Table~\ref{tab:model_benchmarking} reports independent signal firings rather than ensemble decisions.






\bibliographystyle{acl_natbib}
\bibliography{rocling2026}

@inproceedings{xue-etal-2021-mt5,
    title = "m{T}5: A Massively Multilingual Pre-trained Text-to-Text Transformer",
    author = "Xue, Linting  and
      Constant, Noah  and
      Roberts, Adam  and
      Kale, Mihir  and
      Al-Rfou, Rami  and
      Siddhant, Aditya  and
      Barua, Aditya  and
      Raffel, Colin",
    editor = "Toutanova, Kristina  and
      Rumshisky, Anna  and
      Zettlemoyer, Luke  and
      Hakkani-Tur, Dilek  and
      Beltagy, Iz  and
      Bethard, Steven  and
      Cotterell, Ryan  and
      Chakraborty, Tanmoy  and
      Zhou, Yichao",
    booktitle = "Proceedings of the 2021 Conference of the North American Chapter of the Association for Computational Linguistics: Human Language Technologies",
    month = jun,
    year = "2021",
    address = "Online",
    publisher = "Association for Computational Linguistics",
    url = "https://aclanthology.org/2021.naacl-main.41/",
    doi = "10.18653/v1/2021.naacl-main.41",
    pages = "483--498"
}

@article{fan2021beyond,
    title = "Beyond {E}nglish-Centric Multilingual Machine Translation",
    author = "Fan, Angela and Bhosale, Shruti and Schwenk, Holger and Ma, Zhiyi and El-Kishky, Ahmed and Goyal, Siddharth and Baines, Mandeep and Celebi, Onur and Wenzek, Guillaume and Chaudhary, Vishrav and others",
    journal = "Journal of Machine Learning Research",
    volume = "22",
    number = "107",
    pages = "1--48",
    year = "2021",
}

@inproceedings{heffernan-etal-2022-bitext,
    title = "Bitext Mining Using Distilled Sentence Representations for Low-Resource Languages",
    author = "Heffernan, Kevin  and
      {\c{C}}elebi, Onur  and
      Schwenk, Holger",
    editor = "Goldberg, Yoav  and
      Kozareva, Zornitsa  and
      Zhang, Yue",
    booktitle = "Findings of the Association for Computational Linguistics: EMNLP 2022",
    month = dec,
    year = "2022",
    address = "Abu Dhabi, United Arab Emirates",
    publisher = "Association for Computational Linguistics",
    url = "https://aclanthology.org/2022.findings-emnlp.154/",
    doi = "10.18653/v1/2022.findings-emnlp.154",
    pages = "2101--2112"
}

@inproceedings{rei-etal-2020-comet,
    title = "{COMET}: A Neural Framework for {MT} Evaluation",
    author = "Rei, Ricardo  and
      Stewart, Craig  and
      Farinha, Ana C  and
      Lavie, Alon",
    editor = "Webber, Bonnie  and
      Cohn, Trevor  and
      He, Yulan  and
      Liu, Yang",
    booktitle = "Proceedings of the 2020 Conference on Empirical Methods in Natural Language Processing (EMNLP)",
    month = nov,
    year = "2020",
    address = "Online",
    publisher = "Association for Computational Linguistics",
    url = "https://aclanthology.org/2020.emnlp-main.213/",
    doi = "10.18653/v1/2020.emnlp-main.213",
    pages = "2685--2702"
}

@article{artetxe-schwenk-2019-massively,
    title = "Massively Multilingual Sentence Embeddings for Zero-Shot Cross-Lingual Transfer and Beyond",
    author = "Artetxe, Mikel  and
      Schwenk, Holger",
    editor = "Lee, Lillian  and
      Johnson, Mark  and
      Roark, Brian  and
      Nenkova, Ani",
    journal = "Transactions of the Association for Computational Linguistics",
    volume = "7",
    year = "2019",
    address = "Cambridge, MA",
    publisher = "MIT Press",
    url = "https://aclanthology.org/Q19-1038/",
    doi = "10.1162/tacl_a_00288",
    pages = "597--610"
}

@inproceedings{he2021debertav3,
    title = "{DeBERTaV3}: Improving {DeBERTa} using {ELECTRA}-Style Pre-Training with Gradient-Disentangled Embedding Sharing",
    author = "He, Pengcheng and Gao, Jianfeng and Chen, Weizhu",
    booktitle = "The Eleventh International Conference on Learning Representations",
    year = "2023",
    url = "https://openreview.net/forum?id=sE7-XhLxHA",
}

@misc{spacy,
    title = "{spaCy}: Industrial-strength Natural Language Processing in {P}ython",
    author = "Honnibal, Matthew and Montani, Ines",
    year = "2017",
    publisher = "Zenodo",
    doi = "10.5281/zenodo.1212303",
}

@book{wordnet,
  title = {WordNet: An Electronic Lexical Database},
  editor = {Christiane Fellbaum},
  year = {1998},
  publisher = {MIT Press}
}

@inproceedings{ranathunga-etal-2024-quality,
    title = "Quality Does Matter: A Detailed Look at the Quality and Utility of Web-Mined Parallel Corpora",
    author = "Ranathunga, Surangika  and
      De Silva, Nisansa  and
      Menan, Velayuthan  and
      Fernando, Aloka  and
      Rathnayake, Charitha",
    editor = "Graham, Yvette  and
      Purver, Matthew",
    booktitle = "Proceedings of the 18th Conference of the European Chapter of the Association for Computational Linguistics (Volume 1: Long Papers)",
    month = mar,
    year = "2024",
    address = "St. Julian{'}s, Malta",
    publisher = "Association for Computational Linguistics",
    url = "https://aclanthology.org/2024.eacl-long.52",
    pages = "860--880",
}

@article{nllb2022,
  title = {No Language Left Behind: Scaling Human-Centered Machine Translation},
  author = "{NLLB Team} and Costa-juss{\`a}, Marta R. and Cross, James and {\c{C}}elebi, Onur and others",
  journal = {arXiv preprint arXiv:2207.04672},
  year = {2022},
  url = {https://arxiv.org/abs/2207.04672}
}

@article{ji2023survey,
    title = "Survey of Hallucination in Natural Language Generation",
    author = "Ji, Ziwei and Lee, Nayeon and Frieske, Rita and Yu, Tiezheng and Su, Dan and Xu, Yan and Ishii, Etsuko and Bang, Ye Jin and Madotto, Andrea and Fung, Pascale",
    journal = "{ACM} Computing Surveys",
    volume = "55",
    number = "12",
    pages = "1--38",
    year = "2023",
    publisher = "Association for Computing Machinery",
    doi = "10.1145/3571730",
}

@inproceedings{dale-etal-2023-halomi,
    title = "{H}al{O}mi: A Manually Annotated Benchmark for Multilingual Hallucination and Omission Detection in Machine Translation",
    author = {Dale, David  and
      Voita, Elena  and
      Lam, Janice  and
      Hansanti, Prangthip  and
      Ropers, Christophe  and
      Kalbassi, Elahe  and
      Gao, Cynthia  and
      Barrault, Lo{\"i}c  and
      Costa-juss{\`a}, Marta R.},
    editor = "Bouamor, Houda  and
      Pino, Juan  and
      Bali, Kalika",
    booktitle = "Proceedings of the 2023 Conference on Empirical Methods in Natural Language Processing",
    month = dec,
    year = "2023",
    address = "Singapore",
    publisher = "Association for Computational Linguistics",
    url = "https://aclanthology.org/2023.emnlp-main.42/",
    doi = "10.18653/v1/2023.emnlp-main.42",
    pages = "638--653"
}

@inproceedings{lee2018hallucinations,
  title     = {Hallucinations in Neural Machine Translation},
  author    = {Lee, Katherine and Firat, Orhan and Agarwal, Ashish and Fannjiang, Clara and Sussillo, David},
  booktitle = {NeurIPS 2018 Workshop on Interpretability and Robustness in Audio, Speech, and Language},
  year      = {2018},
  address   = {Montr{\'e}al, Canada},
  url       = {https://openreview.net/pdf?id=SkxJ-309FQ}
}

@inproceedings{gal2016dropout,
    title = "Dropout as a {B}ayesian Approximation: Representing Model Uncertainty in Deep Learning",
    author = "Gal, Yarin and Ghahramani, Zoubin",
    booktitle = "Proceedings of the 33rd International Conference on Machine Learning",
    series = "Proceedings of Machine Learning Research",
    volume = "48",
    pages = "1050--1059",
    year = "2016",
    publisher = "PMLR",
}

@inproceedings{kim-etal-2017-predictor,
    title = "Predictor-Estimator using Multilevel Task Learning with Stack Propagation for Neural Quality Estimation",
    author = "Kim, Hyun  and
      Lee, Jong-Hyeok  and
      Na, Seung-Hoon",
    editor = "Bojar, Ond{\v{r}}ej  and
      Buck, Christian  and
      Chatterjee, Rajen  and
      Federmann, Christian  and
      Graham, Yvette  and
      Haddow, Barry  and
      Huck, Matthias  and
      Yepes, Antonio Jimeno  and
      Koehn, Philipp  and
      Kreutzer, Julia",
    booktitle = "Proceedings of the Second Conference on Machine Translation",
    month = sep,
    year = "2017",
    address = "Copenhagen, Denmark",
    publisher = "Association for Computational Linguistics",
    url = "https://aclanthology.org/W17-4763/",
    doi = "10.18653/v1/W17-4763",
    pages = "562--568"
}

@inproceedings{vaswani2017attention,
    title = "Attention Is All You Need",
    author = "Vaswani, Ashish and Shazeer, Noam and Parmar, Niki and Uszkoreit, Jakob and Jones, Llion and Gomez, Aidan N. and Kaiser, {\L}ukasz and Polosukhin, Illia",
    booktitle = "Advances in Neural Information Processing Systems",
    volume = "30",
    year = "2017",
}

@inproceedings{dale-etal-2023-detecting,
    title = "Detecting and Mitigating Hallucinations in Machine Translation: Model Internal Workings Alone Do Well, Sentence Similarity Even Better",
    author = "Dale, David and Voita, Elena and Barrault, Lo{\"i}c and Costa-juss{\`a}, Marta R.",
    editor = "Rogers, Anna  and
      Boyd-Graber, Jordan  and
      Okazaki, Naoaki",
    booktitle = "Proceedings of the 61st Annual Meeting of the Association for Computational Linguistics (Volume 1: Long Papers)",
    month = jul,
    year = "2023",
    address = "Toronto, Canada",
    publisher = "Association for Computational Linguistics",
    url = "https://aclanthology.org/2023.acl-long.3/",
    doi = "10.18653/v1/2023.acl-long.3",
    pages = "36--50"
}

@inproceedings{ferrando-etal-2022-towards,
    title = "Towards Opening the Black Box of Neural Machine Translation: Source and Target Interpretations of the Transformer",
    author = "Ferrando, Javier  and
      G{\'a}llego, Gerard I.  and
      Alastruey, Belen  and
      Escolano, Carlos  and
      Costa-juss{\`a}, Marta R.",
    editor = "Goldberg, Yoav  and
      Kozareva, Zornitsa  and
      Zhang, Yue",
    booktitle = "Proceedings of the 2022 Conference on Empirical Methods in Natural Language Processing",
    month = dec,
    year = "2022",
    address = "Abu Dhabi, United Arab Emirates",
    publisher = "Association for Computational Linguistics",
    url = "https://aclanthology.org/2022.emnlp-main.599/",
    doi = "10.18653/v1/2022.emnlp-main.599",
    pages = "8756--8769"
}

@inproceedings{guerreiro-etal-2023-looking,
    title = "Looking for a Needle in a Haystack: A Comprehensive Study of Hallucinations in Neural Machine Translation",
    author = "Guerreiro, Nuno M.  and
      Voita, Elena  and
      Martins, Andr{\'e}",
    editor = "Vlachos, Andreas  and
      Augenstein, Isabelle",
    booktitle = "Proceedings of the 17th Conference of the European Chapter of the Association for Computational Linguistics",
    month = may,
    year = "2023",
    address = "Dubrovnik, Croatia",
    publisher = "Association for Computational Linguistics",
    url = "https://aclanthology.org/2023.eacl-main.75/",
    doi = "10.18653/v1/2023.eacl-main.75",
    pages = "1059--1075"
}

@inproceedings{zhou-etal-2021-detecting,
    title = "Detecting Hallucinated Content in Conditional Neural Sequence Generation",
    author = "Zhou, Chunting  and
      Neubig, Graham  and
      Gu, Jiatao  and
      Diab, Mona  and
      Guzm{\'a}n, Francisco  and
      Zettlemoyer, Luke  and
      Ghazvininejad, Marjan",
    editor = "Zong, Chengqing  and
      Xia, Fei  and
      Li, Wenjie  and
      Navigli, Roberto",
    booktitle = "Findings of the Association for Computational Linguistics: ACL-IJCNLP 2021",
    month = aug,
    year = "2021",
    address = "Online",
    publisher = "Association for Computational Linguistics",
    url = "https://aclanthology.org/2021.findings-acl.120/",
    doi = "10.18653/v1/2021.findings-acl.120",
    pages = "1393--1404"
}

@inproceedings{zhou-etal-2019-density,
    title = "Density Matching for Bilingual Word Embedding",
    author = "Zhou, Chunting  and
      Ma, Xuezhe  and
      Wang, Di  and
      Neubig, Graham",
    editor = "Burstein, Jill  and
      Doran, Christy  and
      Solorio, Thamar",
    booktitle = "Proceedings of the 2019 Conference of the North {A}merican Chapter of the Association for Computational Linguistics: Human Language Technologies, Volume 1 (Long and Short Papers)",
    month = jun,
    year = "2019",
    address = "Minneapolis, Minnesota",
    publisher = "Association for Computational Linguistics",
    url = "https://aclanthology.org/N19-1161/",
    doi = "10.18653/v1/N19-1161",
    pages = "1588--1598"
}

@article{de2019survey,
    title = "Survey on Publicly Available {S}inhala Natural Language Processing Tools and Research",
    author = "De Silva, Nisansa",
    journal = "arXiv preprint arXiv:1906.02358",
    year = "2026",
}

@inproceedings{lewis-etal-2020-bart,
    title = "{BART}: Denoising Sequence-to-Sequence Pre-training for Natural Language Generation, Translation, and Comprehension",
    author = "Lewis, Mike  and
      Liu, Yinhan  and
      Goyal, Naman  and
      Ghazvininejad, Marjan  and
      Mohamed, Abdelrahman  and
      Levy, Omer  and
      Stoyanov, Veselin  and
      Zettlemoyer, Luke",
    editor = "Jurafsky, Dan  and
      Chai, Joyce  and
      Schluter, Natalie  and
      Tetreault, Joel",
    booktitle = "Proceedings of the 58th Annual Meeting of the Association for Computational Linguistics",
    month = jul,
    year = "2020",
    address = "Online",
    publisher = "Association for Computational Linguistics",
    url = "https://aclanthology.org/2020.acl-main.703/",
    doi = "10.18653/v1/2020.acl-main.703",
    pages = "7871--7880"
}

@inproceedings{wang-sennrich-2020-exposure,
    title = "On Exposure Bias, Hallucination and Domain Shift in Neural Machine Translation",
    author = "Wang, Chaojun  and
      Sennrich, Rico",
    editor = "Jurafsky, Dan  and
      Chai, Joyce  and
      Schluter, Natalie  and
      Tetreault, Joel",
    booktitle = "Proceedings of the 58th Annual Meeting of the Association for Computational Linguistics",
    month = jul,
    year = "2020",
    address = "Online",
    publisher = "Association for Computational Linguistics",
    url = "https://aclanthology.org/2020.acl-main.326/",
    doi = "10.18653/v1/2020.acl-main.326",
    pages = "3544--3552"
}

@inproceedings{raunak-etal-2021-curious,
    title = "The Curious Case of Hallucinations in Neural Machine Translation",
    author = "Raunak, Vikas  and
      Menezes, Arul  and
      Junczys-Dowmunt, Marcin",
    editor = "Toutanova, Kristina  and
      Rumshisky, Anna  and
      Zettlemoyer, Luke  and
      Hakkani-Tur, Dilek  and
      Beltagy, Iz  and
      Bethard, Steven  and
      Cotterell, Ryan  and
      Chakraborty, Tanmoy  and
      Zhou, Yichao",
    booktitle = "Proceedings of the 2021 Conference of the North American Chapter of the Association for Computational Linguistics: Human Language Technologies",
    month = jun,
    year = "2021",
    address = "Online",
    publisher = "Association for Computational Linguistics",
    url = "https://aclanthology.org/2021.naacl-main.92/",
    doi = "10.18653/v1/2021.naacl-main.92",
    pages = "1172--1183"
}

@article{fomicheva2020unsupervised,
  title={Unsupervised quality estimation for neural machine translation},
  author={Fomicheva, Marina and Sun, Shuo and Yankovskaya, Lisa and Blain, Fr{\'e}d{\'e}ric and Guzm{\'a}n, Francisco and Fishel, Mark and Aletras, Nikolaos and Chaudhary, Vishrav and Specia, Lucia},
  journal={Transactions of the Association for Computational Linguistics},
  volume={8},
  pages={539--555},
  year={2020},
  publisher={MIT Press One Rogers Street, Cambridge, MA 02142-1209, USA journals-info~…}
}

@inproceedings{papineni-etal-2002-bleu,
    title = "{B}leu: a Method for Automatic Evaluation of Machine Translation",
    author = "Papineni, Kishore  and
      Roukos, Salim  and
      Ward, Todd  and
      Zhu, Wei-Jing",
    editor = "Isabelle, Pierre  and
      Charniak, Eugene  and
      Lin, Dekang",
    booktitle = "Proceedings of the 40th Annual Meeting of the Association for Computational Linguistics",
    month = jul,
    year = "2002",
    address = "Philadelphia, Pennsylvania, USA",
    publisher = "Association for Computational Linguistics",
    url = "https://aclanthology.org/P02-1040/",
    doi = "10.3115/1073083.1073135",
    pages = "311--318"
}

@inproceedings{martindale-etal-2019-identifying,
    title = "Identifying Fluently Inadequate Output in Neural and Statistical Machine Translation",
    author = "Martindale, Marianna  and
      Carpuat, Marine  and
      Duh, Kevin  and
      McNamee, Paul",
    editor = "Forcada, Mikel  and
      Way, Andy  and
      Haddow, Barry  and
      Sennrich, Rico",
    booktitle = "Proceedings of Machine Translation Summit XVII: Research Track",
    month = aug,
    year = "2019",
    address = "Dublin, Ireland",
    publisher = "European Association for Machine Translation",
    url = "https://aclanthology.org/W19-6623/",
    pages = "233--243"
}

@inproceedings{jalili-sabet-etal-2020-simalign,
    title = "{S}im{A}lign: High Quality Word Alignments Without Parallel Training Data Using Static and Contextualized Embeddings",
    author = {Jalili Sabet, Masoud  and
      Dufter, Philipp  and
      Yvon, Fran{\c{c}}ois  and
      Sch{\"u}tze, Hinrich},
    editor = "Cohn, Trevor  and
      He, Yulan  and
      Liu, Yang",
    booktitle = "Findings of the Association for Computational Linguistics: EMNLP 2020",
    month = nov,
    year = "2020",
    address = "Online",
    publisher = "Association for Computational Linguistics",
    url = "https://aclanthology.org/2020.findings-emnlp.147/",
    doi = "10.18653/v1/2020.findings-emnlp.147",
    pages = "1627--1643"
}

@inproceedings{zhang2019bertscore,
    title = "{BERTScore}: Evaluating Text Generation with {BERT}",
    author = "Zhang, Tianyi and Kishore, Varsha and Wu, Felix and Weinberger, Kilian Q. and Artzi, Yoav",
    booktitle = "International Conference on Learning Representations",
    year = "2020",
    url = "https://openreview.net/forum?id=SkeHuCVFDr",
}

@inproceedings{feng-etal-2022-language,
    title = "Language-agnostic {BERT} Sentence Embedding",
    author = "Feng, Fangxiaoyu  and
      Yang, Yinfei  and
      Cer, Daniel  and
      Arivazhagan, Naveen  and
      Wang, Wei",
    editor = "Muresan, Smaranda  and
      Nakov, Preslav  and
      Villavicencio, Aline",
    booktitle = "Proceedings of the 60th Annual Meeting of the Association for Computational Linguistics (Volume 1: Long Papers)",
    month = may,
    year = "2022",
    address = "Dublin, Ireland",
    publisher = "Association for Computational Linguistics",
    url = "https://aclanthology.org/2022.acl-long.62/",
    doi = "10.18653/v1/2022.acl-long.62",
    pages = "878--891"
}

@inproceedings{conneau-etal-2020-unsupervised,
    title = "Unsupervised Cross-lingual Representation Learning at Scale",
    author = "Conneau, Alexis  and
      Khandelwal, Kartikay  and
      Goyal, Naman  and
      Chaudhary, Vishrav  and
      Wenzek, Guillaume  and
      Guzm{\'a}n, Francisco  and
      Grave, Edouard  and
      Ott, Myle  and
      Zettlemoyer, Luke  and
      Stoyanov, Veselin",
    editor = "Jurafsky, Dan  and
      Chai, Joyce  and
      Schluter, Natalie  and
      Tetreault, Joel",
    booktitle = "Proceedings of the 58th Annual Meeting of the Association for Computational Linguistics",
    month = jul,
    year = "2020",
    address = "Online",
    publisher = "Association for Computational Linguistics",
    url = "https://aclanthology.org/2020.acl-main.747/",
    doi = "10.18653/v1/2020.acl-main.747",
    pages = "8440--8451"
}

@article{liu-etal-2020-multilingual,
    title = "Multilingual Denoising Pre-training for Neural Machine Translation",
    author = "Liu, Yinhan and Gu, Jiatao and Goyal, Naman and Li, Xian and Edunov, Sergey and Ghazvininejad, Marjan and Lewis, Mike and Zettlemoyer, Luke",
    journal = "Transactions of the Association for Computational Linguistics",
    volume = "8",
    year = "2020",
    address = "Cambridge, MA",
    publisher = "MIT Press",
    url = "https://aclanthology.org/2020.tacl-1.47/",
    doi = "10.1162/tacl_a_00343",
    pages = "726--742",
}

@inproceedings{poliak-etal-2018-hypothesis,
    title = "Hypothesis Only Baselines in Natural Language Inference",
    author = "Poliak, Adam and Naradowsky, Jason and Haldar, Aparajita and Rudinger, Rachel and Van Durme, Benjamin",
    booktitle = "Proceedings of the Seventh Joint Conference on Lexical and Computational Semantics",
    month = jun,
    year = "2018",
    address = "New Orleans, Louisiana",
    publisher = "Association for Computational Linguistics",
    url = "https://aclanthology.org/S18-2023/",
    doi = "10.18653/v1/S18-2023",
    pages = "180--191",
}

@inproceedings{gururangan-etal-2018-annotation,
    title = "Annotation Artifacts in Natural Language Inference Data",
    author = "Gururangan, Suchin and Swayamdipta, Swabha and Levy, Omer and Schwartz, Roy and Bowman, Samuel R. and Smith, Noah A.",
    booktitle = "Proceedings of the 2018 Conference of the North {A}merican Chapter of the Association for Computational Linguistics: Human Language Technologies, Volume 2 (Short Papers)",
    month = jun,
    year = "2018",
    address = "New Orleans, Louisiana",
    publisher = "Association for Computational Linguistics",
    url = "https://aclanthology.org/N18-2017/",
    doi = "10.18653/v1/N18-2017",
    pages = "107--112",
}

\end{document}